\documentclass[conference,letterpaper]{IEEEtran}
\IEEEoverridecommandlockouts
\usepackage[dvips]{graphicx} 
\usepackage{amsmath} 

\usepackage{multirow}
\usepackage[left=0.71in,top=0.94in,right=0.71in,bottom=1.18in]{geometry}
\usepackage{algorithm}
\usepackage{algorithmicx}
\usepackage{algpseudocode}

\usepackage{subcaption}
\usepackage{color} 

\usepackage{siunitx}

\usepackage{soul}
\usepackage{todonotes}

\graphicspath{{figures/}}

\global\long\def\cluster{C} 
\global\long\def\stats{\boldsymbol{\mu}} 
\global\long\def\vel{\boldsymbol{v}}
\global\long\def\corner{\boldsymbol{c}}
\global\long\def\flow{\boldsymbol{f}}

\begin{document}
\title{\huge Independent Motion Detection with Event-driven Cameras}

\author{V. Vasco\textsuperscript{1}, A. Glover\textsuperscript{1}, E. Mueggler\textsuperscript{2}, D. Scaramuzza\textsuperscript{2}, L. Natale\textsuperscript{1} and C. Bartolozzi\textsuperscript{1}\\%
\textit{\textsuperscript{1}iCub Facility, Istituto Italiano di Tecnologia, Genova, Italy}\\
\vspace{0.1cm}
\textit{\{valentina.vasco, arren.glover, lorenzo.natale, chiara.bartolozzi\}@iit.it}\\
\textit{\textsuperscript{2}Robotics and Perception Group, University of Zurich, Zurich, Switzerland}\\
\textit{\{mueggler, sdavide\}@ifi.uzh.ch}
}


\maketitle
\begin{abstract}

Unlike standard cameras that send intensity images at a constant frame rate, event-driven cameras asynchronously report pixel-level brightness changes, offering low latency and high temporal resolution (both in the order of micro-seconds). As such, they have great potential for fast and low power vision algorithms for robots. 
Visual tracking, for example, is easily achieved even for very fast stimuli, as only moving objects cause brightness changes. However, cameras mounted on a moving robot are typically non-stationary and the same tracking problem becomes confounded by background clutter events due to the robot ego-motion. In this paper, we propose a method for segmenting the motion of an independently moving object for event-driven cameras.    
Our method detects and tracks corners in the event stream and learns the statistics of their motion as a function of the robot's joint velocities when no independently moving objects are present.
During robot operation, independently moving objects are identified by discrepancies between the predicted corner velocities from ego-motion and the measured corner velocities.
We validate the algorithm on data collected from the neuromorphic iCub robot. We achieve a precision of $\sim90\%$ and show that the method is robust to changes in speed of both the head and the target.

\end{abstract}

\section{Introduction}


The direct interaction between a robot and its surroundings is one of the major challenges in robotics. The iCub, designed to be anthropomorphic with a three-and-a-half year old child, primarily uses vision to measure the state of the external environment and, as such, visual motion estimation is fundamental.

Unlike standard cameras that read the full sensor array to produce images at a fixed frame rate, event cameras only report change in pixel-level brightness above a threshold. The ``events'' are produced independently and asynchronously for each pixel sensor. They offer a high dynamic range (\SI{140}{\decibel} as compared to \SI{60}{\decibel} of standard cameras), together with low latency and high temporal resolution (both in the order of \emph{micro}-seconds). Processing of redundant information is avoided as pixels that do not experience a change simply do not activate.

As such, event cameras are a promising technology for fast, accurate and low power vision algorithms for robots in dynamic environments. Since the output of event cameras is fundamentally different from standard cameras (a continuous, asynchronous stream of events instead of a sequence of images), new algorithms are required to deal with these data. The iCub~\cite{Metta10nn} is a humanoid robot designed with sensory and actuation capabilities to interact with a dynamic environment, in which objects and the robot itself move simultaneously. The neuromorphic iCub is equipped with a stereo pair of ATIS event cameras~\cite{Posch11ssc}.

Solving a problem such as visual tracking can be performed almost trivially using a stationary event camera: only the motion of the target causes events to be produced. The problem of segmenting the background from the target is inherently solved by the sensor. However, the movement of an event camera mounted on a robot causes events to be generated due to all contrast in the field-of-view and the problem of tracking or recognising the motion of a target object becomes more difficult. Differentiating the event camera signal caused by ego-motion from the signal caused by independent motion has many potential uses in event-driven robotics.


The problem of Independent Motion Detection (IMD) was first studied in~\cite{Costeira98ijcv,Irani98pami}. However, as noted by~\cite{Roberts09cvpr}, the apparent motion induced by the ego-motion has ``some degree of statistical regularity'' and ``the structure of the environment [..] is far from arbitrary'' for many applications (in their case, autonomous ground vehicles). Also constraints on the vehicle motion were suggest to be exploited~\cite{Sabzevari16tro}.

Besides visual sensing, most robots are also equipped with proprioceptive sensors such as inertial measurement units and joint encoders. Thus, a typical strategy is to use the knowledge of the robot kinematics to predict the apparent motion induced by ego-motion~\cite{Vieville95ras,Fanello13icra,Kumar15humanoids}. However, due to imprecision in the mechanical structure and sensor acquisition errors, significant noise is present in these predictions. Therefore, \cite{Fanello13icra} proposed to \emph{learn} the correlation joint velocities and flow statistics, without requiring a kinematic model of the robot. Independent motion is found by predicting a probability distribution of the optical flow and identifying flow vectors that do not belong to this distribution. This approach estimates ego-motion correctly, even if the majority of the image plane is independent motion (e.g., a large object in front of the camera), which is not possible by using methods relying on vision alone.


In this paper we present a method to segment events caused by ego-motion from those caused by independent object motion. Such a technique has wide applicability for event-driven algorithms in robotics. For example, identifying independent object motion makes detection and tracking of moving objects simple. In a dynamic environment moving objects are usually very relevant to behaviour (e.g. for avoidance or attentional interaction with human collaborators handling objects).
 Alternatively, segmentation of ego-motion events can be used to remove outliers to improve event-based visual odometry methods (e.g.~\cite{Kim16eccv, Rebecq17ral}) in dynamic scenes. The use of event cameras for this task is driven by the strong potential for low-latency, low-power robotic vision. Our algorithm takes advantage of previous work in event-based corner detection~\cite{Vasco16iros} as well as traditional approaches to ego-motion segmentation~\cite{Fanello13icra,Kumar15humanoids}.






\section{Related Work} \label{sec:related_work}


\subsection{Independent Motion Detection for Robot Applications}

In \cite{Ciliberto11iros}, the authors proposed the \textit{MotionCUT} framework, based on Lucas-Kanade tracker failures. They observed that Lucas-Kanade algorithm typically fails around rotations and occlusions, which are likely caused by objects independently moving. This approach assumes visual motion to be dominated by ego-motion and only a small portion of it to be produced by independent motion. In case of large objects or objects close to the camera, the camera translation component is not negligible and it can produce a failure in the Lucas-Kanade tracking, as well as an independent moving object. The assumption of the ego-motion being the dominant motion of the scene is often violated in dynamic environments where many objects move at the same time.   

In~\cite{Ciliberto12iros}, the kinematics of the robots were assumed to be known, but with input-dependent noise due to mechanical imperfections and sensor noise. To address these issues, the correlation between predicted and tracked features was learned. For the prediction, the depth of the features was estimated using stereo. The statistics were learned in situations without independently moving objects, i.e., the apparent motion was only due to the ego-motion. In the testing phase, these statistics are used to detect anomalies that correspond to independent motion. 

In~\cite{Kumar15humanoids}, the detection of independent motion was used for object segmentation. It builds upon~\cite{Fanello13icra}, but additionally includes inertial measurements.

\subsection{Event-based Motion Estimation}

Much work on motion estimation with event cameras has focused on SLAM and visual odometry algorithms. 
Estimating only the rotational component of a camera has been performed using various methods~\cite{Cook11ijcnn, Kim14bmvc, Gallego17ral}, however, the motion of the iCub's eyes includes a non-negligible amount of translational motion, and these algorithms are not suitable, especially when objects are close to the camera. 

Event-based 6-DOF SLAM and visual odometry methods~\cite{Kim16eccv,Rebecq17ral} assume static scenes and sufficiently large independently moving objects introduce errors. In this paper, we consider environments in which objects are also moving and are interested in their motion, rather than solely the robot's ego-motion. It was shown that specific algorithmic adjustments were required to perform tracking in a highly-dynamic environment, which had simultaneous camera and object motion~\cite{Glover2016iros}. When using an event-based camera, the problem difficulty increases dramatically. In addition, the experiments relied solely on visual data as they did not have access to the robot kinematics. Previously, combining event camera vision with external sensors has been performed using a gyroscope for event-stabilisation~\cite{Delbruck14iscas}.  Again, this method can only estimate the apparent motion due to camera rotation, but not translation.

To segment ego-motion induced events from those induced from an independently moving object, first a measure of the optical flow is required. Optical flow can be easily calculated using event cameras, by fitting planes in the three-dimensional spatio-temporal space in which events exist~\cite{Benosman2014tnnls}. However, similarly to frame-based cameras, when considering only small spatial windows of events, aperture problems cause incorrect flow estimation along long uniform edges. With event-based cameras, the locally spatial plane fitting method, in combination with the temporally asynchronous nature of events, makes it difficult to apply global correction methods.

Unique features are typically used to avoid the aperture problem, from which the true optical flow can be calculated. Event-based corner detection methods have been proposed~\cite{Clady2015nn,Vasco16iros}. In this paper, we use~\cite{Vasco16iros}, that adapted the Harris method to the event data stream. It has also been shown that tracking corners in event-space is possible, resulting in a faster than frame-rate update of corner positions.



\section{Methodology} \label{sec:method}

\begin{figure*}
\centering
\includegraphics[width=0.75\textwidth]{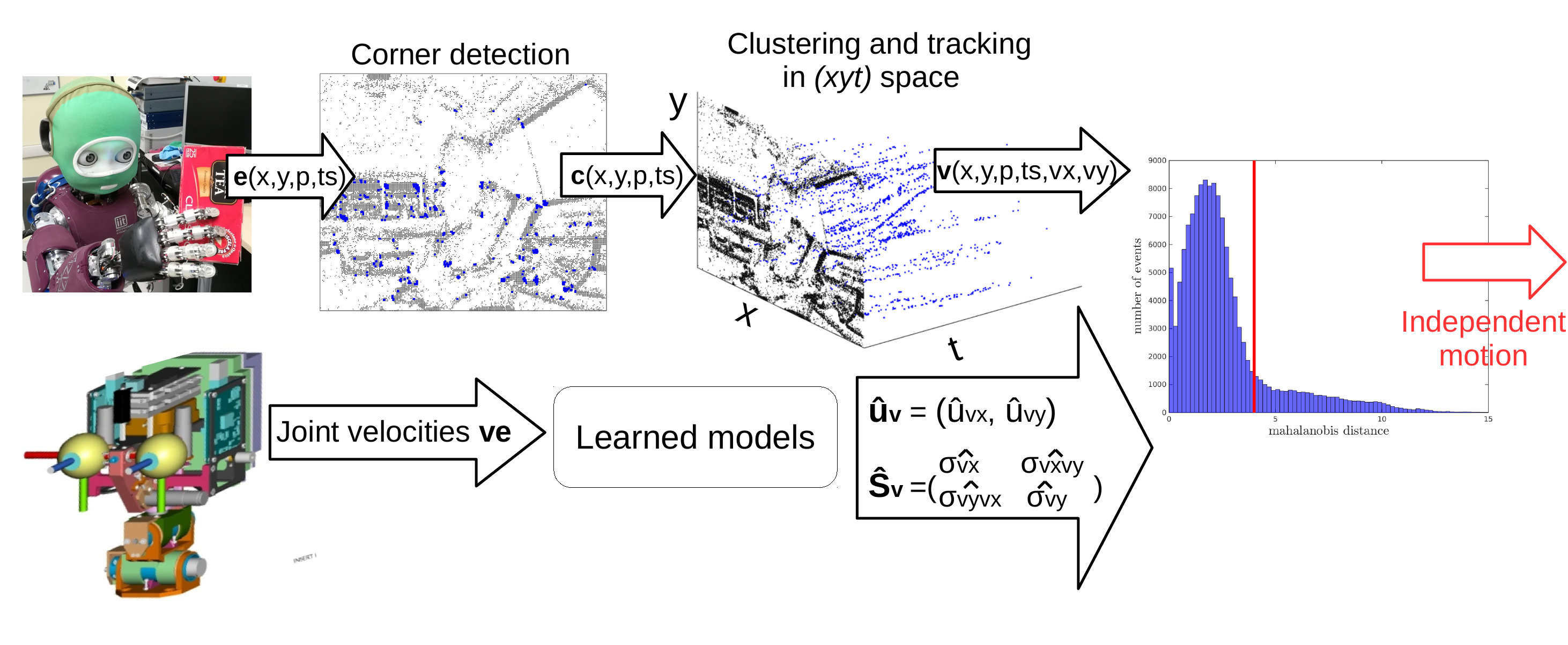}
\caption{Pipeline of the algorithm.}
\label{fig:algorithm}
\end{figure*}

The pipeline for detecting independent motion consists of visual corner detection, tracking and velocity estimation in parallel with the estimation of the joint velocities from motor encoders, as shown in Fig.~\ref{fig:algorithm}. In a learning phase using a completely static scene, a model of the correlation between motor velocities and the resulting visual motion is developed (following~\cite{Fanello13icra,Kumar15humanoids}). During operation, computed visual motion is compared to the expected visual motion given the model and the movement of the robot. Large discrepancies between estimated motion and computed motion can be classified as being caused by an independently moving object.

\subsection{Corner Detection}

The dynamic vision circuitry of the ATIS camera provides a stream of events, in the form of $\{x_i, y_i, p_i, ts_i\}$ (pixel position in $(x, y)$, polarity, and timestamp). The stream of events is first reduced to the subspace of corner events, detected using~\cite{Vasco16iros}, with the following implementation changes. Previously a fixed event window was used that stored a fixed number of the most-recent events globally across the sensor space in the form of a ``surface''~\cite{Benosman2014tnnls}. Good results were achieved with the size of the window tuned according to the scene complexity. However this can be problematic when motion is not uniform over the visual scene (e.g. in our case of an object moving with motion independent of the camera motion), as higher velocities produce a higher number of events per second, manifesting in the event window as ``thicker edges''. Instead of a single global event window, we implemented the same fixed event surface on a local scale, one for each pixel location. The window size becomes no-longer dependent on the particular scene and object motion, but on the feature type used: in our case, corners. As the features are constant, the window size can also be constant and independent of the scene. For corner detection we set the window size to be $2 \times l$, where $l$ is the radius of the window used.



\subsection{Corner clustering and tracking}

Events that are labelled as corners are clustered into corner tracks, from which the velocity of each corner can be estimated. The first cluster is initialized with the first corner event and following corner events are added based on the spatial distance to the clusters: events are added if the distance is less than a threshold $D$; if no cluster is less than $D$ pixels to the current event, a new cluster is created.

Such a greedy corner allocation method can be applied as corner detection error is typically limited to 2 pixels~\cite{Vasco16iros} and we can assume to observe the full trajectory (pixel by pixel) of objects in the event space (i.e. an observed object cannot jump more than one pixel when using an event-based camera, whereas it is a common occurrence for objects that move faster than the frame-rate of a traditional camera).




Each cluster is updated according to a first-in first-out rule: when the maximum size $S$ is reached, the oldest corner event is removed from the cluster. To avoid tracking corner events that do not reflect the current motion, clusters are deleted if they are not updated for a time higher than $t_{refresh}$. 

Corners positions are tracked over time using regression to fit a line in the $(x, y, t)$ space. Given a set of corner events $\corner_i = \{x_i, y_i, p_i, ts_i\}$ that belong to the current cluster $\cluster_k$, we find the function $f$ that minimizes the sum of the squared deviations:

\begin{equation}
\label{eq:1}
f : \min \sum_{i = 1}^n (ts_i - f(x_i, y_i, a, b, c))^2.
\end{equation}
$(a, b, c)$ defines the direction of the line and provides the components of the flow in both directions $\vel = (v_{xi}, v_{yi}) = \frac{1}{c}(a, b)$. To minimise the error on velocity estimation, we define a minimum number of events $m$ for the clusters to be informative. 

The corner event $\corner_i$ is augmented with the additional information from the velocity calculation, generating a flow event: $\flow_i$ $\{x_i,y_i,p_i,ts_i,v_{xi},v_{yi}\}$.

The algorithm is detailed in Algorithm \ref{algorithm1}.

\begin{algorithm}
    \caption{Event-based clustering and tracking}\label{algorithm1}

    \begin{algorithmic}
    \Require $\corner_i$ $\{x_i,y_i,p_i,ts_i\}$
    \For {\textit{each} $\corner_i$}
        \State \textit{Compute the distance $\bf{d_{ik}}$ from the set of clusters $\cluster=\{\cluster_1, ..., \cluster_N\}$}
        \If {$\bf{d_{ik}} < D$}
            \State \textit{Assign $\corner_i$ to $\cluster_k$ : $\bf{d_{ik}} = \min\limits_{k = 1, ..., N} \bf{d_{ik}}$}
        \Else
        		\State \textit{Create a new cluster $\cluster_k$}        
        \EndIf  
        \State \textit{Fit a 3D line to the set of corner events in $\cluster_k$}
		\State \textit{Compute $\vel_i = (v_{xi}, v_{yi})^T$ as slope of the line}        
        \If {$size_{of}(\cluster_k) > S$}
        		\State \textit{Remove the oldest corner event}        
        \EndIf
        \For {\textit{each cluster $\cluster_i \in \cluster$}}
        		\If {$(ts - t_i) > t_{refresh}$}
        			\State \textit{Delete $\cluster_i$}
        		\EndIf
        \EndFor    
        \State \textit{Assign $\vel_i$ to $\corner_i$} 
        \State \textit{Define $\flow_i$ $\{x_i,y_i,p_i,ts_i,v_{xi},v_{yi}\}$}
    \EndFor
    \end{algorithmic}
    
\end{algorithm}

Corner clusters are updated and an estimation of visual velocity is calculated asynchronously as events occur, with a higher than microsecond resolution. The flow of the entire scene can be found by querying all event clusters that are active at any point in time. The most recent flow event within the cluster holds the most up-to-date velocity calculation for each corner cluster.

\subsection{Model Learning}

A model of average visual motion given motor encoder velocities is learned from data. Robot joint velocities, $\vel_e$, can be estimated by differentiating encoder positions and applying a filter~\cite{Janabi-Sharifi2000}. Supervised learning is performed every time joint velocities are read such that the input is $\vel_e(t)$ and the learning signal is the optical flow statistics $\stats_v(t)$, $\textbf{S}_v(t)$ computed from corner tracking. $\stats_v(t)$ and $\textbf{S}_v(t)$ are the mean scene velocities ($\mu_{v_x}, \mu_{v_y}$) and the covariance matrix of all active clusters queried at time $t$. Importantly, the model must be learned (once) in a completely static environment such that ego-motion alone contributes to the visual flow. 

Joint velocities are updated every $10~ms$, while the estimated velocity from the corner clusters is asynchronously updated for each event, with an initial latency of $m$ events. When at least half of the clusters reach $m$ events, we read the encoder velocities and update the statistics, considering only clusters that satisfy the requirement. To avoid using just one cluster which would not be informative to learn scene statistics, we also define a minimum number of clusters $n$ that need to be active at the same time. 
For fast speed motion, the speed of the algorithm is limited by the encoder update frequency, but for slower motion, the update is tailored to the dynamics of the scene, and the update is done only when sufficient information is gathered, saving computation and power.
 


Joint velocities and associated cluster velocities are then used as training examples for $\nu$-SVM ~\cite{CC01a} with an RBF kernel, to learn five regressors, namely $(\hat{\mu_{vx}}, \hat{\mu_{vy}}, \hat{\sigma_{vx}}, \hat{\sigma_{vx vy}}, \hat{\sigma_{vy}})$.

The algorithm is detailed in~\ref{algorithm2}. 

\begin{algorithm}
    \caption{Model learning}\label{algorithm2}

    \begin{algorithmic}
    \Require $\bf{f_{i}}$ $\{x_i,y_i,p_i,ts_i,v_{xi},v_{yi}\}$
    \State \textit{Initialize $\stats_v = (\mu_{v_x}, \mu_{v_y})$, $\textbf{S}_v = ({\sigma_{vx}}, {\sigma_{vx vy}}; {\sigma_{vx vy}}, {\sigma_{vy}})$}
    \For {\textit{each} $\bf{f_{i}}$}
    		\If {\textit{$\frac{N}{2}$ clusters have more than $m$ corner events}}
    			\State \textit{Update $\stats_v = (\mu_{v_x}, \mu_{v_y})$ and $\textbf{S}_v = ({\sigma_{vx}}, {\sigma_{vx vy}}; {\sigma_{vx vy}}, {\sigma_{vy}})$}
    			\State \textit{Associate $\stats_v, \textbf{S}_v \leftarrow \vel_e$}
    		\EndIf     
    \EndFor
    \end{algorithmic}
    
\end{algorithm}

\subsection{Independent Motion Classification}

During robot operation we compare the computed velocity of corner events $\vel_i = (v_x, v_y)$ with the expected distribution $\hat{\stats_v} = (\hat{\mu_{vx}}, \hat{\mu_{vy}})$, $\hat{\textbf{S}_v} = (\hat{\sigma_{vx}}, \hat{\sigma_{vx vy}}; \hat{\sigma_{vx vy}}, \hat{\sigma_{vy}})$ predicted using the model, given the joint velocities. The Mahalanobis distance is used as a metric of how likely the calculated velocities belong to the ego-motion distribution:

\begin{equation}
d = \sqrt{(\vel - \hat{\stats})^T \hat{\textbf{S}}^{-1} (\vel - \hat{\stats})}.
\label{eq:metric}
\end{equation}

Classification is performed using a distance threshold $T$, such that flow events $\flow_i$ that are below the threshold can be assumed to be created from the motion of the robot itself, while those that exceed the threshold can be labelled as independent motion events.

\section{Experiments and Results} \label{sec:experiments}


\subsection{Experiments}

We characterized and tested the algorithm on data collected from the event camera (ATIS) mounted on the iCub robot. iCub head has a total of 6 Degrees of Freedom (DoFs): $3$ for the neck (pitch, roll and yaw), and $3$ for the eyes (tilt, version and vergence). The iCub GazeController~\cite{Roncone2016} was used to move the robot head position, controlling both neck and eyes independently. Ego-motion was generated by defining 3D target positions in the environment to gaze at, and different speeds were achieved by specifying the time for the trajectory to be completed.

During the learning phase, in order to get a representative dataset for training, we sampled a static environment selecting targets for the controller distributed in a rectangular area, while changing the times for gazing. The data collected were used to train the ego-motion prediction model.

\begin{figure}
    \centering
\begin{subfigure}[b]{0.13\textwidth} 
        \includegraphics[width=\textwidth]{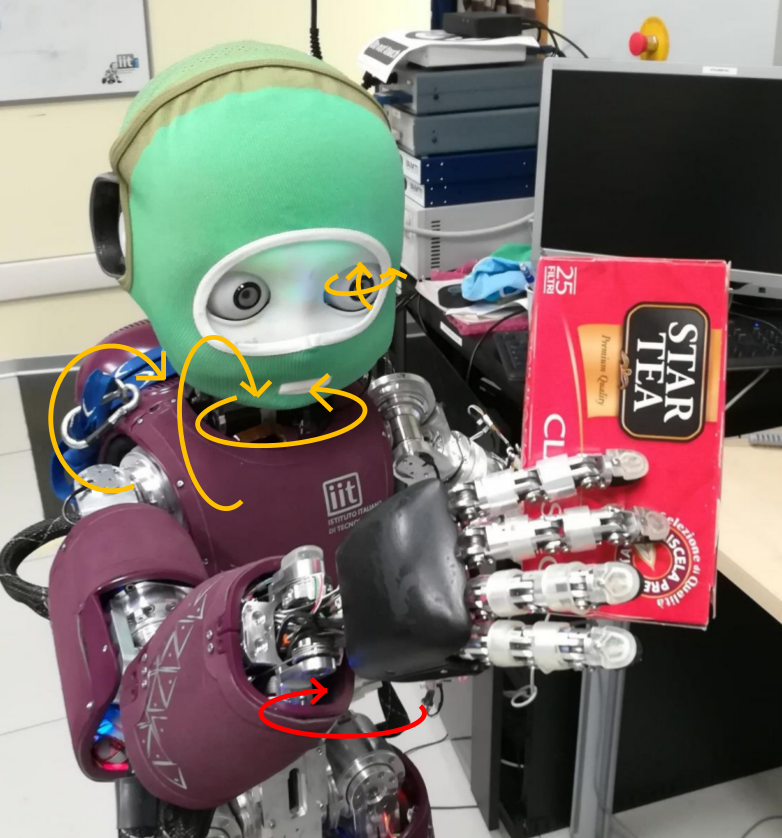}
        \caption{}
        \label{fig:icubfront}
    \end{subfigure} 
    \begin{subfigure}[b]{0.187\textwidth} 
        \includegraphics[width=\textwidth]{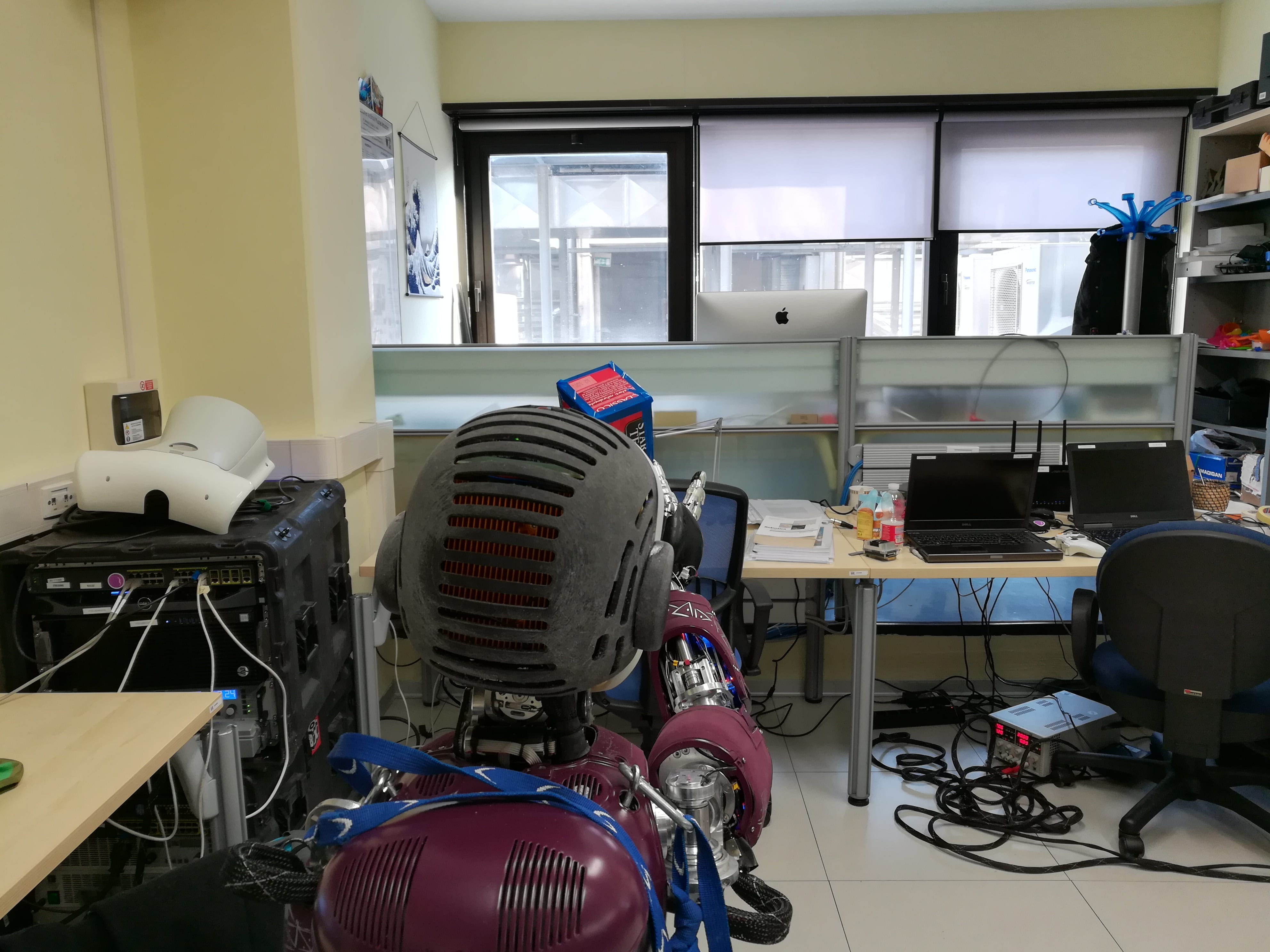}
        \caption{}
        \label{fig:icubback}
    \end{subfigure}  
\caption{The iCub moving its head while grasping the object. (a) shows the DoFs of the head/neck and of the wrist, (b) shows the background environment. The dataset was used for the testing phase.}
\label{fig:setup}
\end{figure}

For the testing phase, we performed two sets of experiments. In the first, we moved the head randomly in the environment at a fixed speed and simultaneously the hand around the yaw axis, while holding an object (a tea box). This method was used to control the velocity of the object and the ground truth. The hand moved at $4$ different speeds, in order to evaluate the detection response with different velocities of the independently moving object. 
In the second experiment, we changed the velocity of the head, while maintaining fixed the velocity of the hand, in order to evaluate the detection response with different velocities of the ego-motion. Fig.~\ref{fig:setup} shows the experimental setup. We controlled the hand and the head velocities, respectively at $120, 130, 140$ and $150^{\circ}/s$ and $3, 5$ and $10^{\circ}/s$.

For all testing datasets, the region-of-interest defining the position of the object that underwent independent motion was  labelled by hand. The corner events falling within the region-of-interest formed the ground-truth true-positive detections.

We empirically selected the following parameters: $l = 5~px,~D = 5~px,~n = 5,~S = 50,~m = 15,~t_{refresh} = 1~s$.

\begin{figure}
\centering
\includegraphics[width=0.3\textwidth]{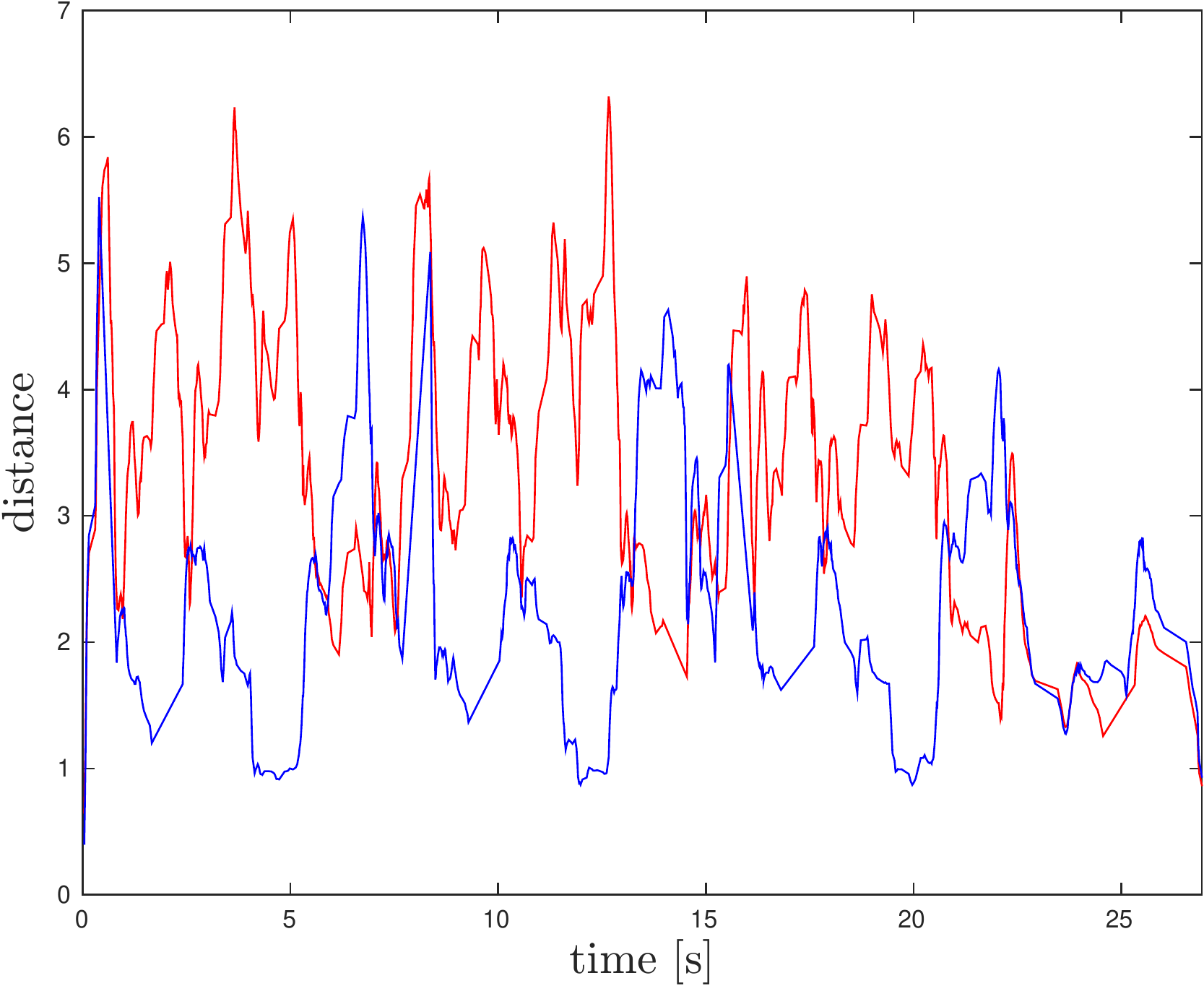}
\caption{The distance between velocity vectors of the predicted motion and velocity due to ego-motion (blue line) and independent motion (red line). The distance metric defines the error between velocity vectors, which, for the independent motion, is more often larger. The dataset used is for a hand speed of $130^{\circ}/s$.}
\label{fig:metric}
\end{figure}

\begin{figure}
    \centering
    \includegraphics[width=0.3\textwidth]{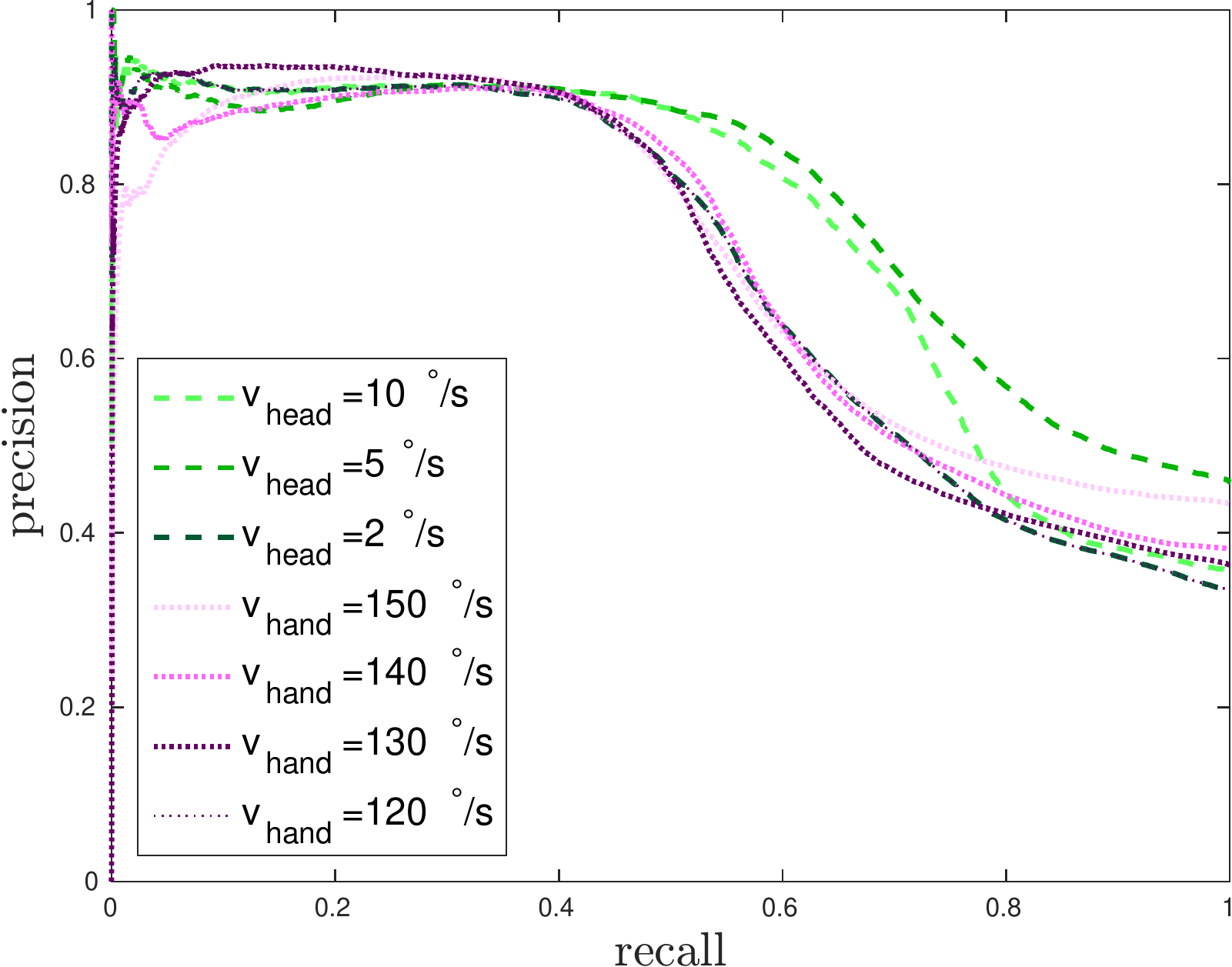}
\caption{Precision and recall curves for different velocities of the hand and head.}
\label{fig:PR}
\end{figure}

\subsection{Results}

Sparse corner flow events were compared to the learned model using the metric defined in Eq.~\ref{eq:metric}. Fig.~\ref{fig:metric} shows the distance between the velocity vectors and the predicted motion computed according to Eq.~\ref{eq:metric}, grouped into background and independent motion (blue and red line) according to the ground truth. The average distance over $\sim10~ms$ for each group is shown. In general, velocity vectors that belong to independent motion exhibit higher distances to the model than background vectors. This indicates the potential for separating ego-motion and independent motion using the proposed method.
At some points in the dataset the distance for the independent motion corners drops to a similar level as the ego-motion corners (red and blue lines overlap, for example, between $0-1~s$, $7-8~s$, $15-16~s$ and in the last $5~s$). This happens as the object stops moving and becomes indistinguishable from the background motion. 
These points do not correspond to failure in the detection algorithm, but represent an intrinsic limitation that originates from the use of motion to detect the target.

The performance of the algorithm, on an event-by-event basis, as the detection threshold $T$ changes, was evaluated in terms of precision and recall, shown in Fig.~\ref{fig:PR}. At low thresholds (i.e. low recall), the precision is $\sim90~\%$, which indicates that a strong ``independent motion'' response was always present in the system. Such a response is caused by noise in the detection algorithm, however a precision of $100~\%$ may not be required for many robotic applications. The precision is stable over a wide range of thresholds, until $\sim40~\%$ recall rate is achieved. We can therefore select a threshold $T$ to achieve a precision of $\sim90~\%$. Performances are consistent with different speeds of the target and iCub head. The algorithm is therefore robust to changes in velocities and a valid threshold can be chosen that should be robust to speed variation.

\begin{figure*}
    \centering
\begin{subfigure}[b]{0.23\textwidth} 
        \includegraphics[width=\textwidth]{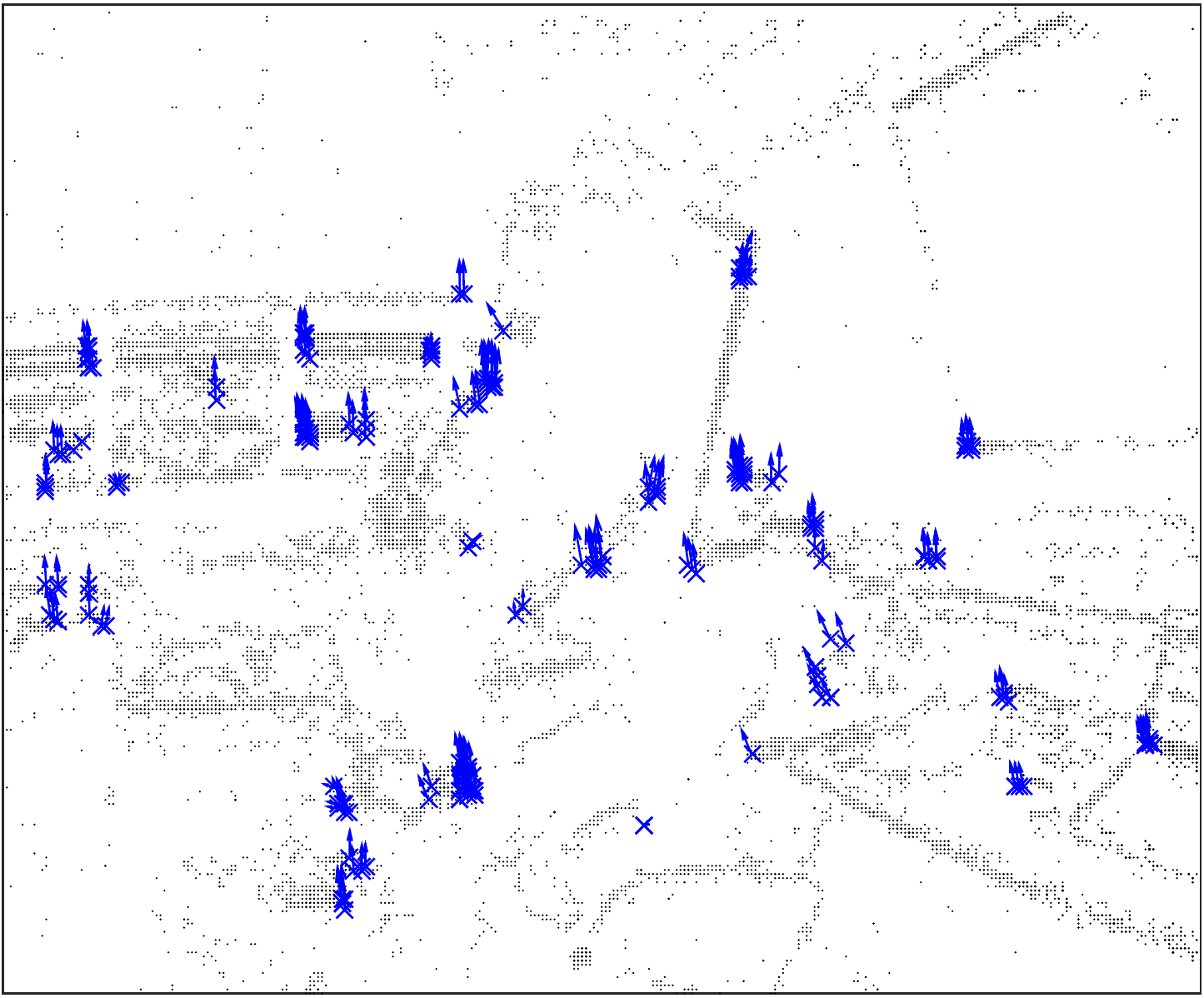}
        \caption{}
        \label{fig:frame1}
    \end{subfigure} 
    \begin{subfigure}[b]{0.23\textwidth} 
        \includegraphics[width=\textwidth]{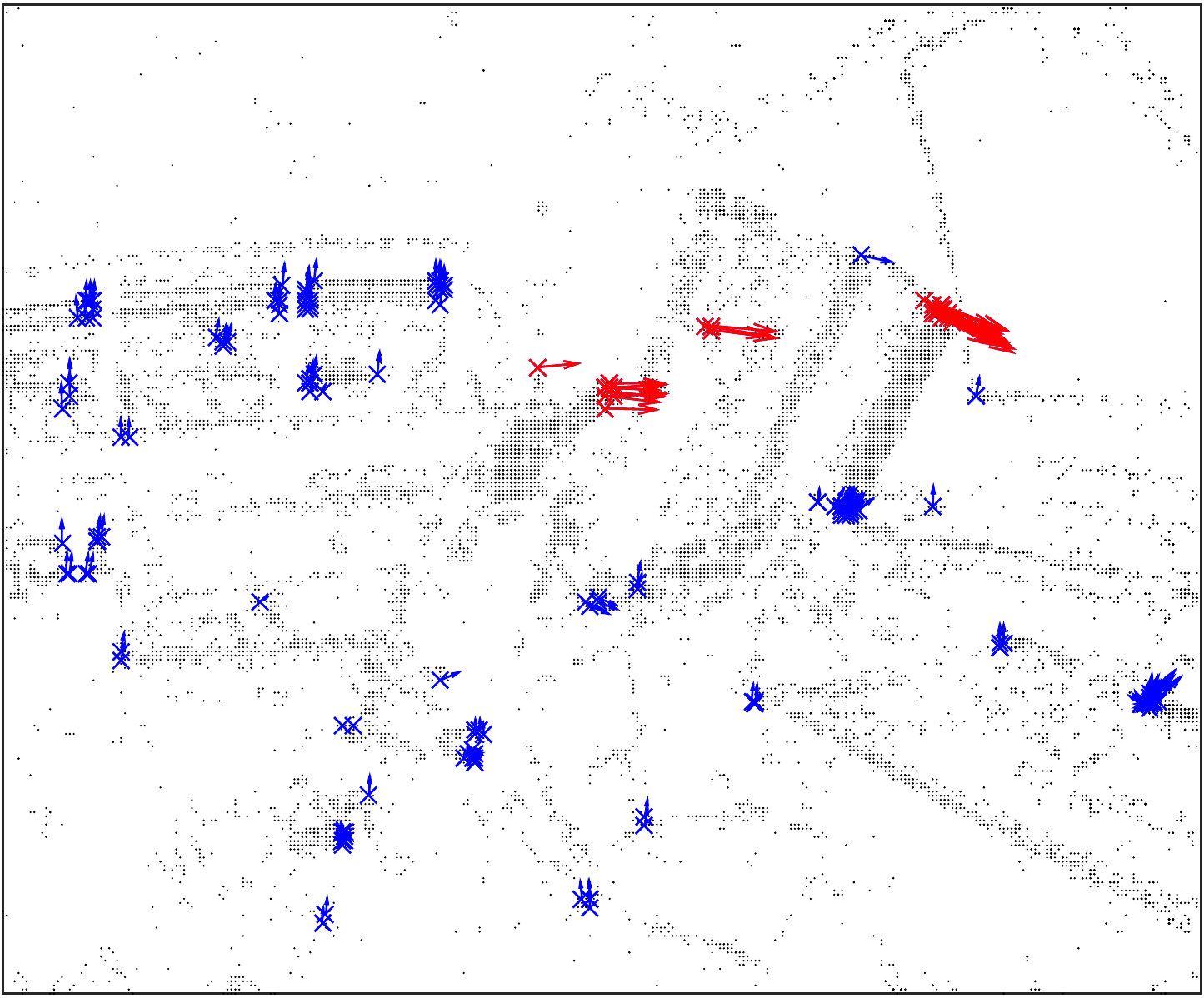}
        \caption{}
        \label{fig:frame2}
    \end{subfigure}  
    \begin{subfigure}[b]{0.23\textwidth} 
        \includegraphics[width=\textwidth]{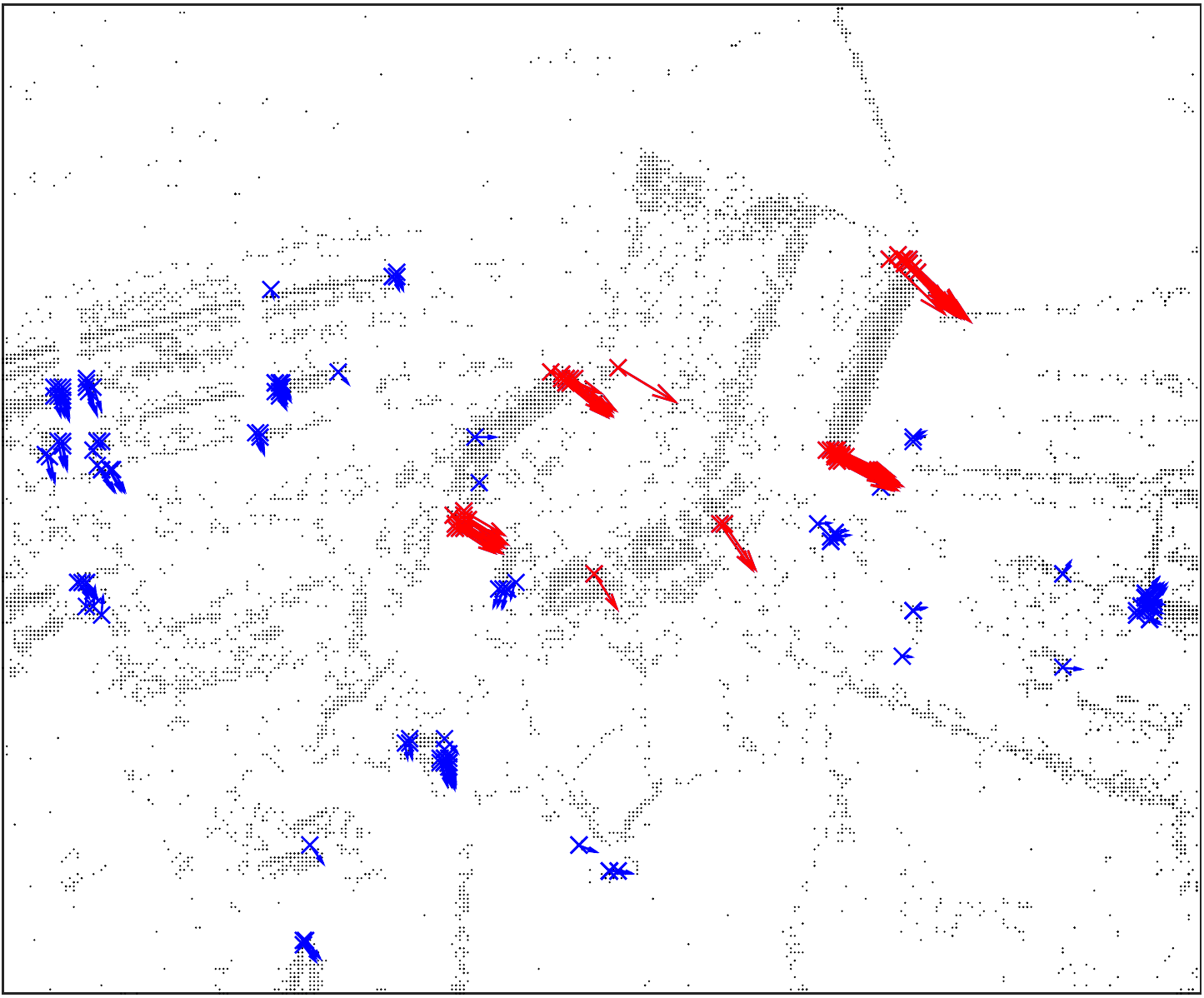}
        \caption{}
        \label{fig:frame3}
    \end{subfigure} \\
    \begin{subfigure}[b]{0.25\textwidth} 
        \includegraphics[width=\textwidth]{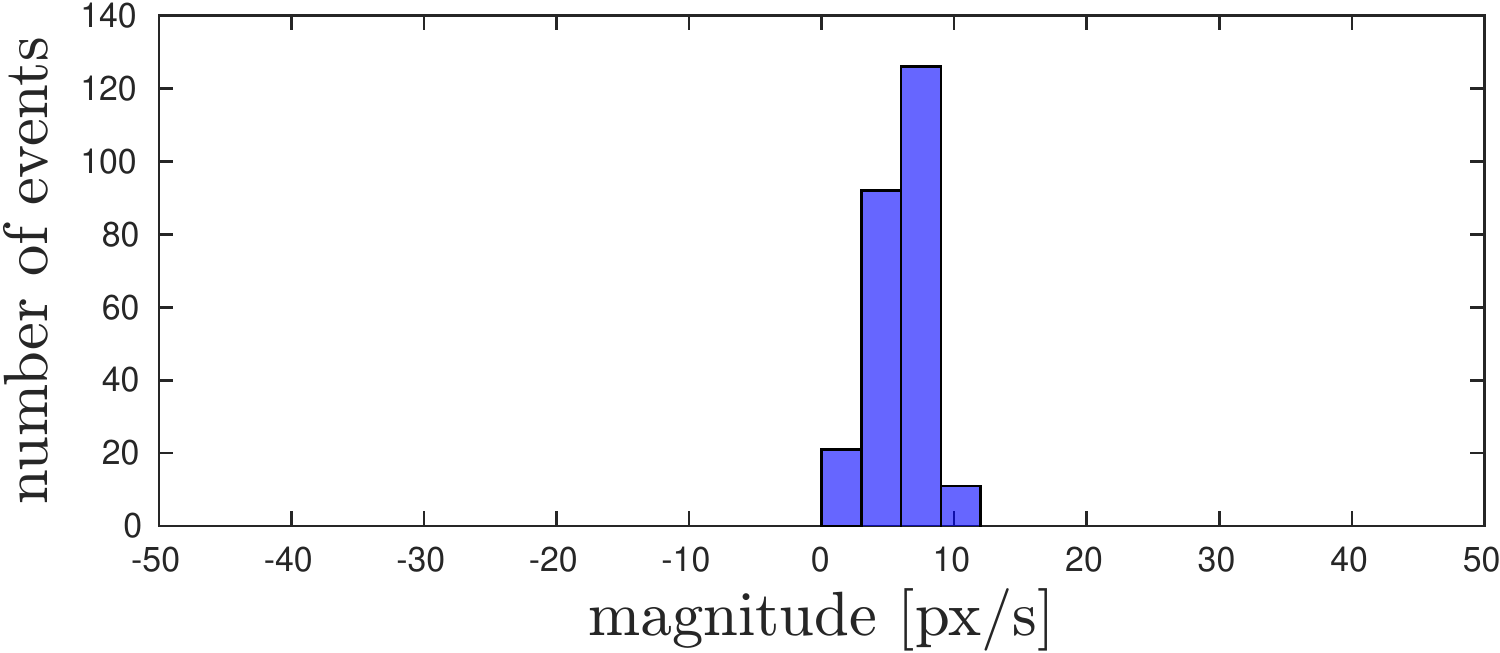}
        \caption{}
        \label{fig:mag1}
    \end{subfigure} 
    \begin{subfigure}[b]{0.25\textwidth} 
        \includegraphics[width=\textwidth]{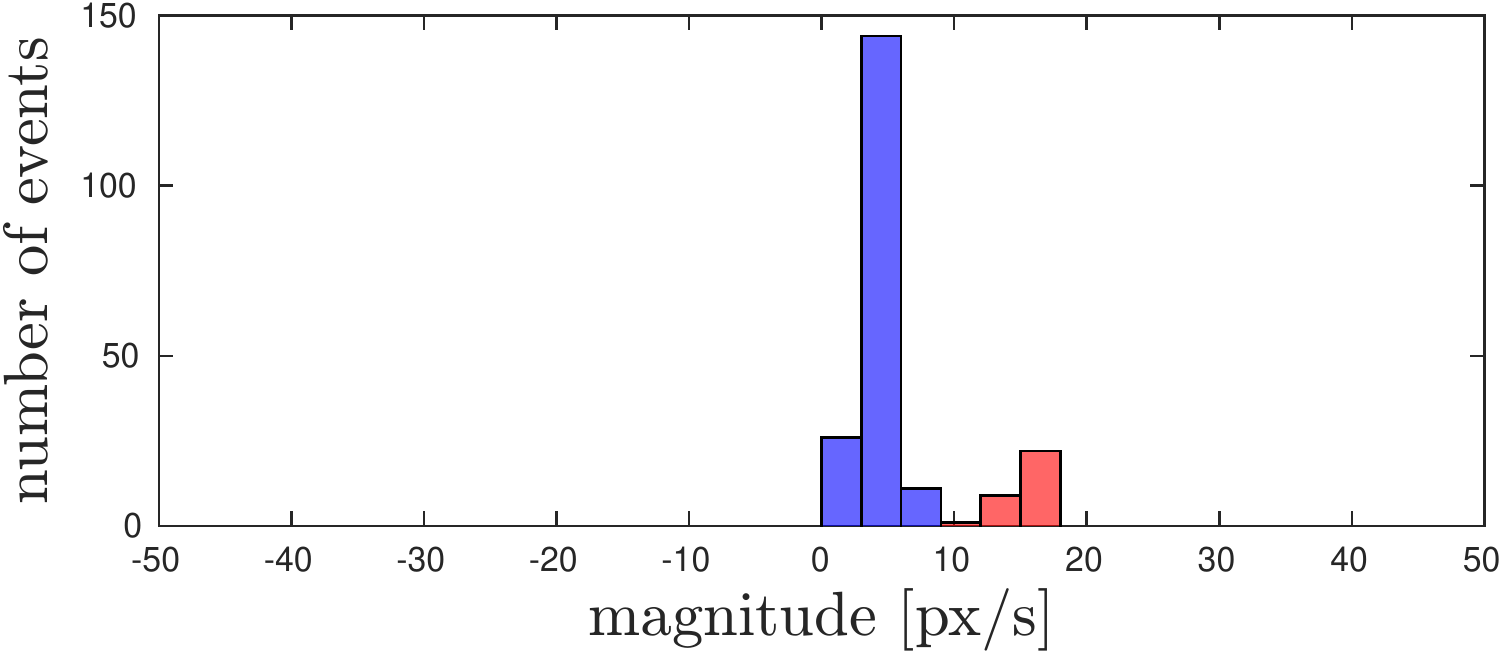}
        \caption{}
        \label{fig:mag2}
    \end{subfigure}  
    \begin{subfigure}[b]{0.25\textwidth} 
        \includegraphics[width=\textwidth]{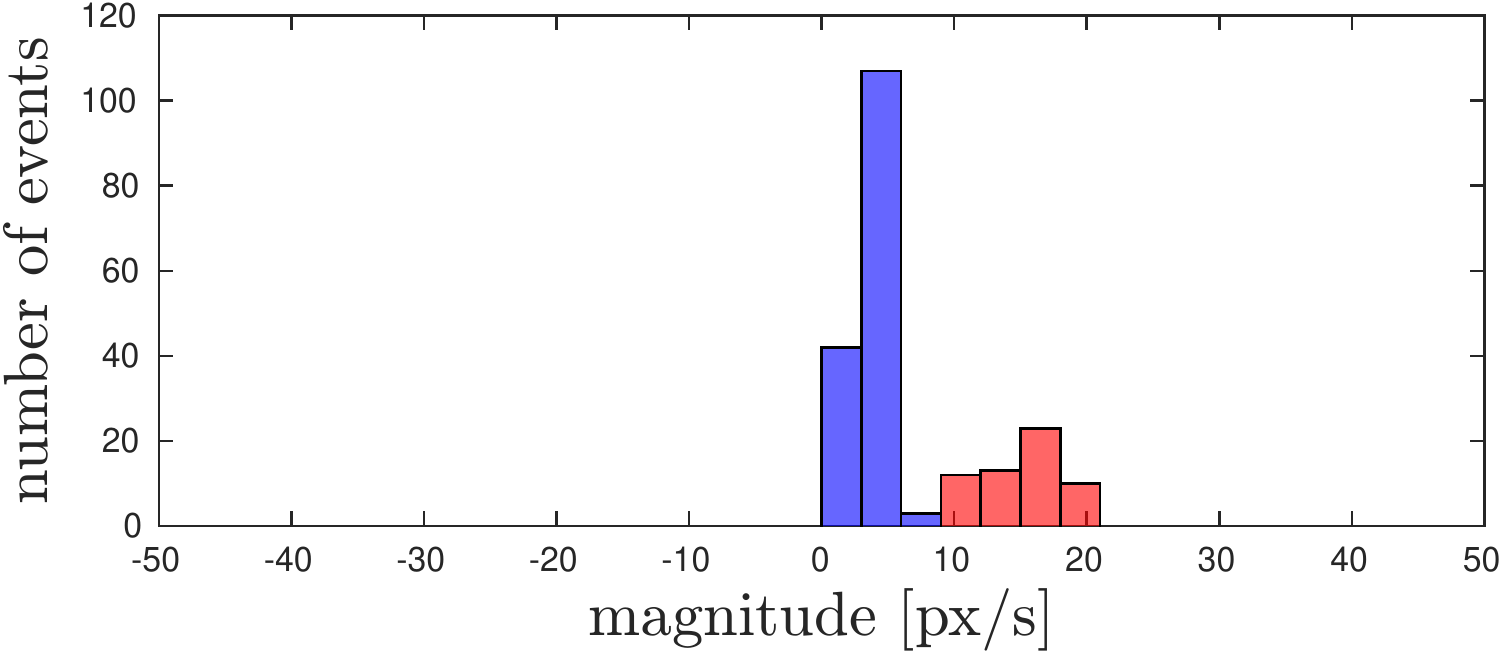}
        \caption{}
        \label{fig:mag3}
     \end{subfigure} \\
        \begin{subfigure}[b]{0.25\textwidth} 
        \includegraphics[width=\textwidth]{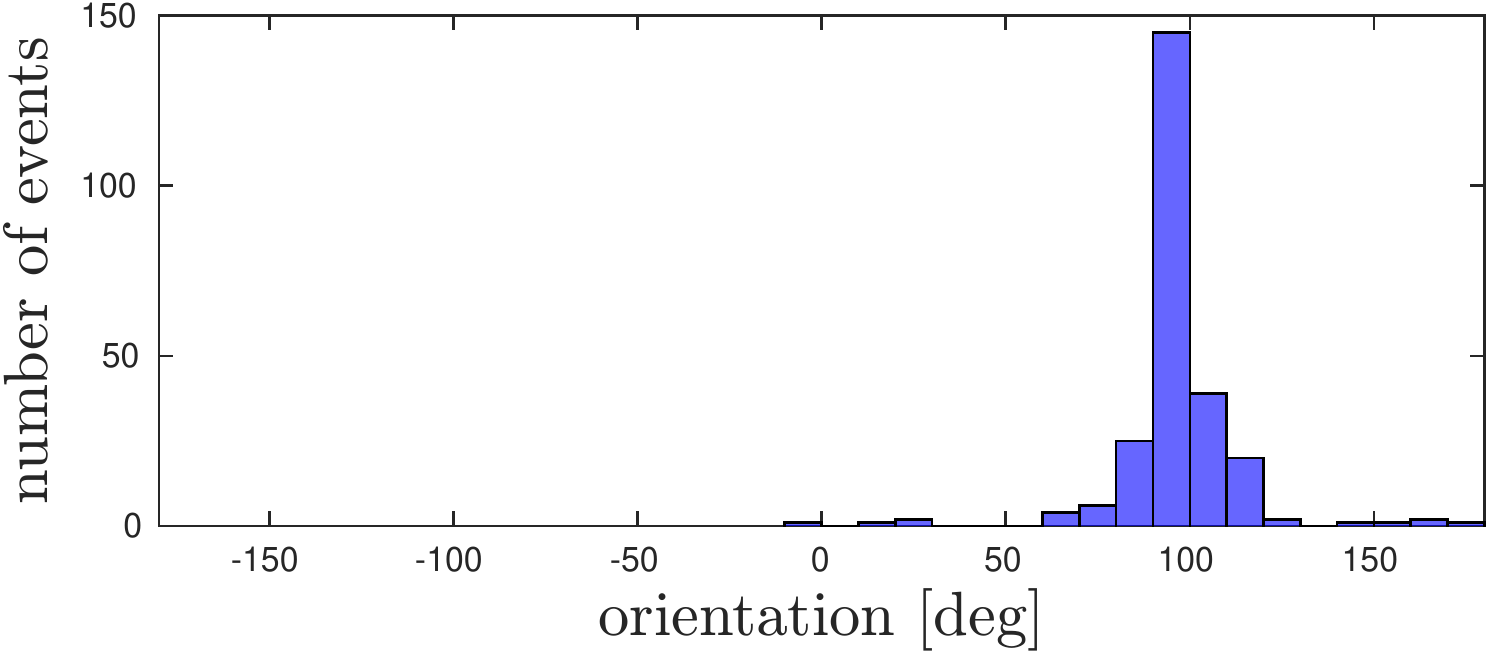}
        \caption{}
        \label{fig:ori1}
    \end{subfigure} 
    \begin{subfigure}[b]{0.25\textwidth} 
        \includegraphics[width=\textwidth]{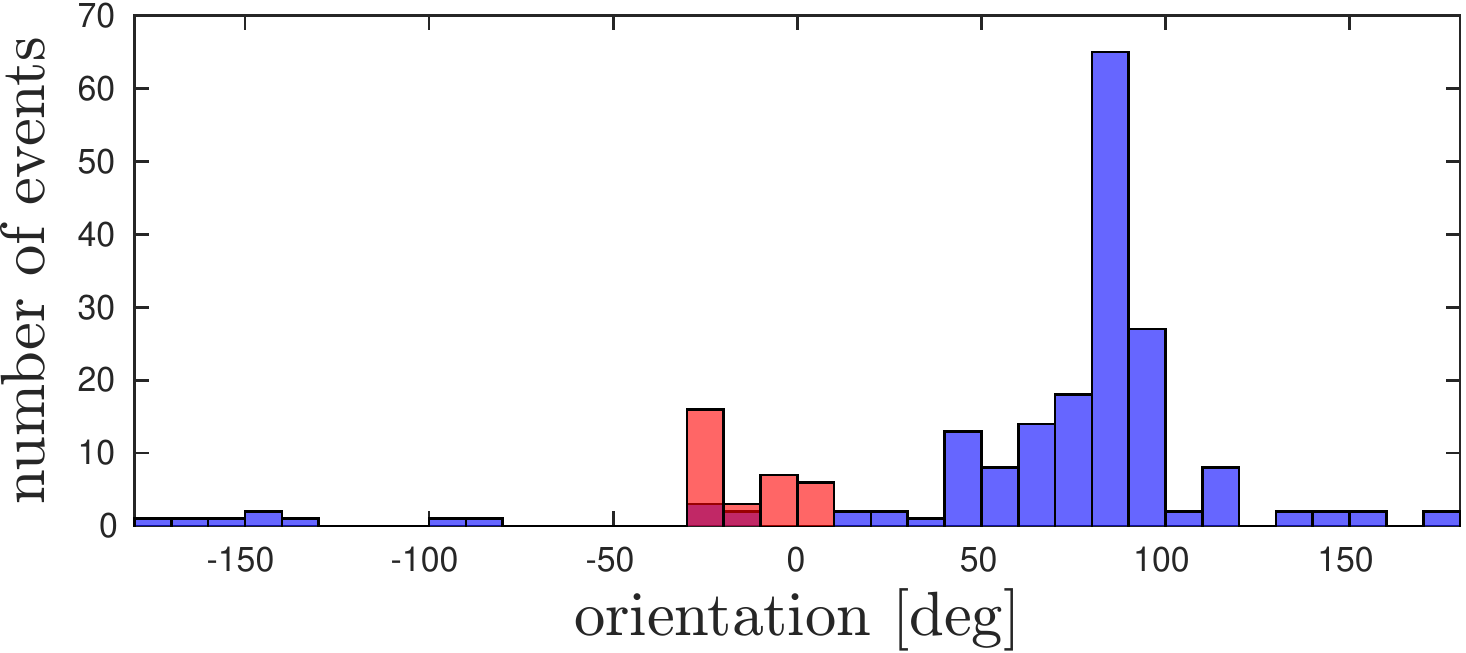}
        \caption{}
        \label{fig:ori2}
    \end{subfigure}  
    \begin{subfigure}[b]{0.25\textwidth} 
        \includegraphics[width=\textwidth]{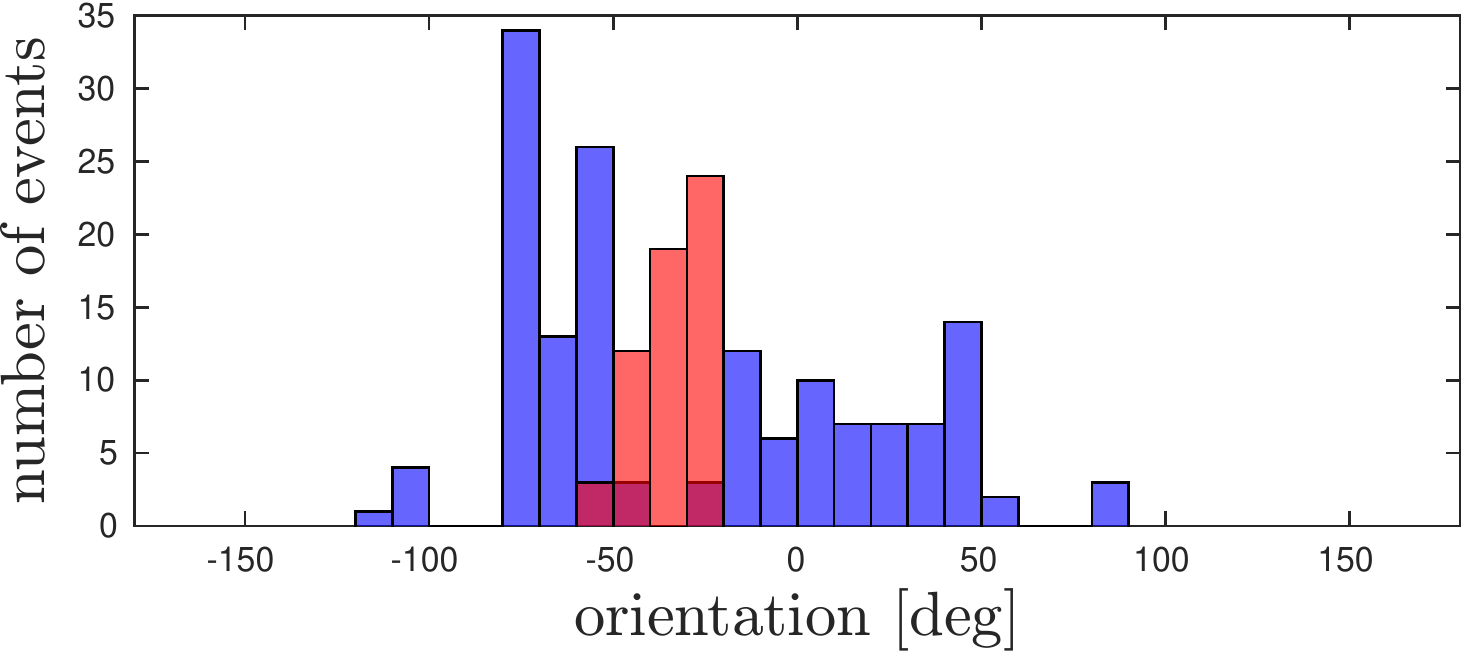}
        \caption{}
        \label{fig:ori3}
    \end{subfigure} 

\caption{Qualitative assessment of independent motion segmentation. (a, b, c) Examples of frames at different time intervals. Events, corner events and independent motion events are shown in black, blue and red respectively, with the relative velocities. 
(d, e, f) Motion magnitude and orientation statistics for the corresponding frames. When the object is steady, all of corner events are classified as ego-motion events (a-d). When the object starts moving, its corners are classified as independent motion (b-e, c-f).}
\label{fig:frames}
\end{figure*}

Example snapshots of events (accumulated over $350~ms$ and labelled according to the selected threshold $T = 4$), are shown in Fig.~\ref{fig:frames} along with the corresponding motion distributions (in orientation and magnitude). In Fig.~\ref{fig:frame1}, corner events on the target fall within the ego-motion distribution, as the target is not moving. Coherently with Fig.~\ref{fig:metric}, we can select a proper threshold to separate the independent motion from the background, both in conditions in which motion magnitudes and orientations are separable, as shown in Fig.~\ref{fig:frames}~(\subref{fig:frame2}\subref{fig:mag2}\subref{fig:ori2}), and when only one of the two components is separable (a different magnitude but orientation falling in the same distribution is shown in Fig.~\ref{fig:frames} (\subref{fig:frame3}\subref{fig:mag3}\subref{fig:ori3}). 

Some corner events are not labelled as independent motion even though they belong to the moving object (Fig~\ref{fig:frames} \subref{fig:frame2}\subref{fig:frame3}). This happens mainly when motion direction changes, in this case, the relative motion between the object and the background approaches zero and independent motion is similar to the ego-motion generated flow. Additionally, during sharp changes in motion, there is a small latency in recovering the clusters and re-evaluating the new velocity. 

\begin{figure}
    \centering
\begin{subfigure}[b]{0.44\textwidth}
        \centering
        \includegraphics[width=0.7\linewidth]{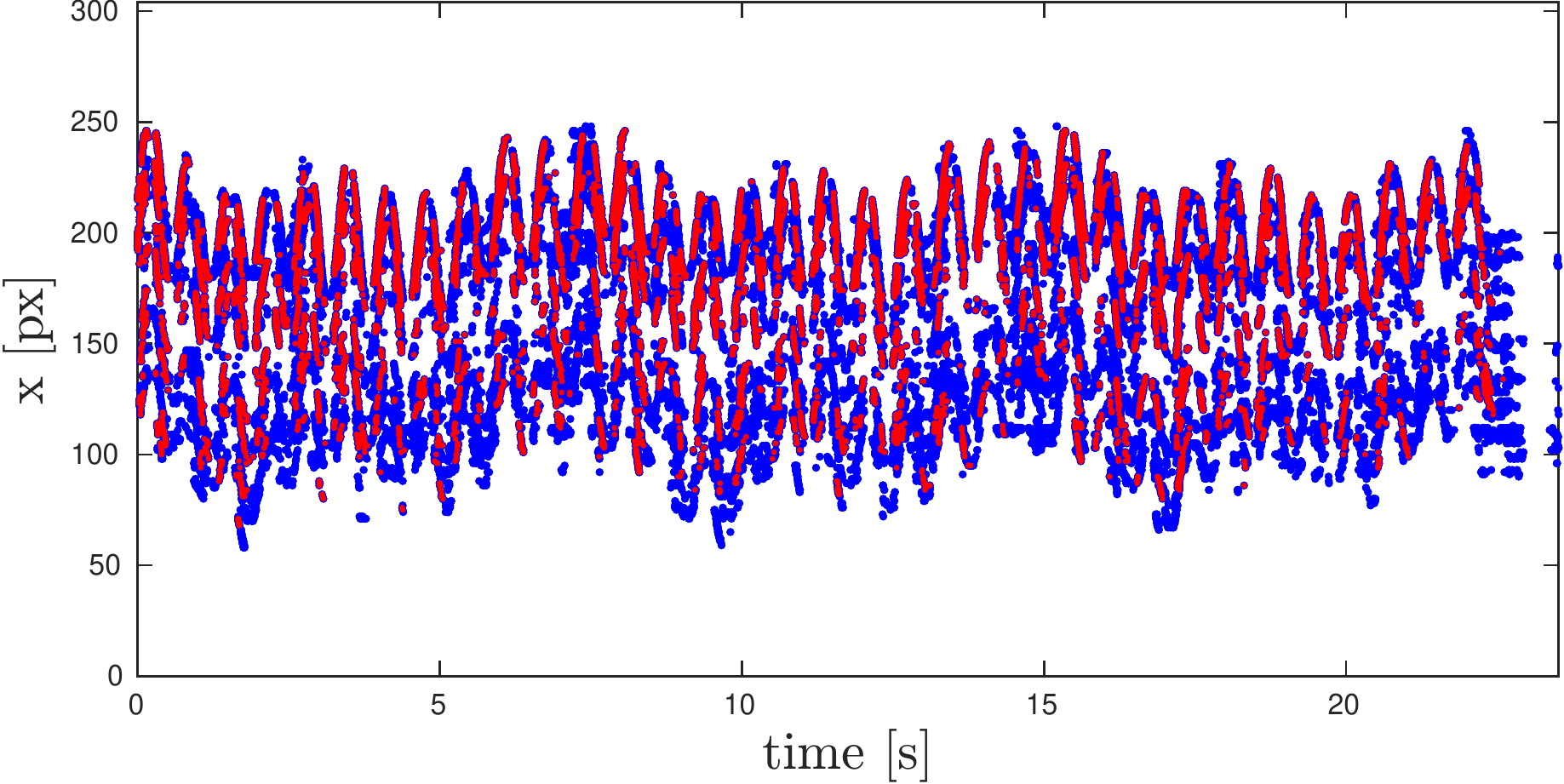}
        \caption{}
        \label{fig:trackX}
    \end{subfigure} \\
    \begin{subfigure}[b]{0.44\textwidth} 
        \centering
        \includegraphics[width=0.7\linewidth]{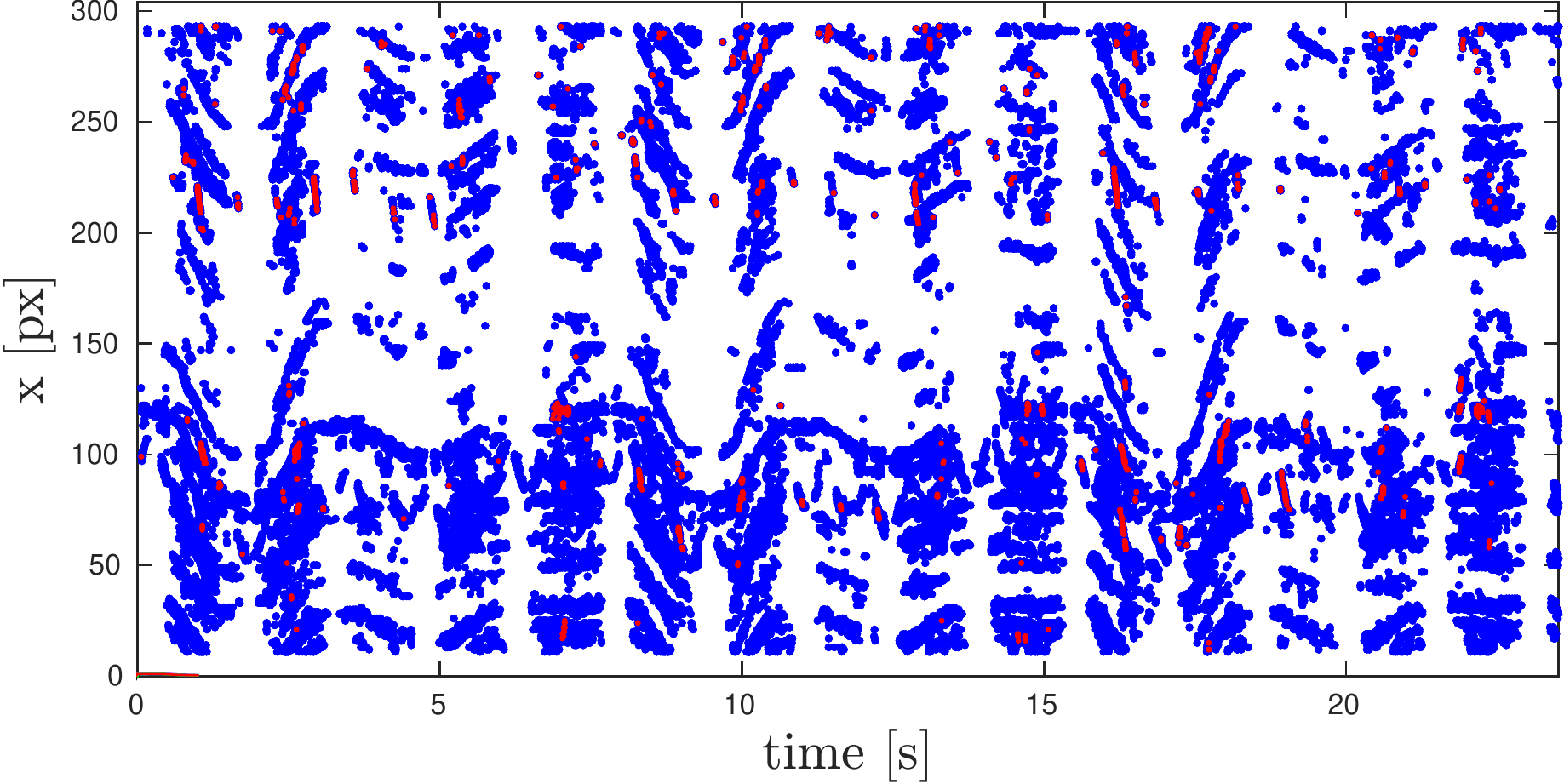}
        \caption{}
        \label{fig:trackY}
    \end{subfigure}  
\caption{Trajectories of the $x$ coordinate traced by events that belong to the object (blue and red for corner and independent motion events) (a), and by events that belong to the background (b). Only one trajectory is shown for visualisation.}
\label{fig:tracking}
\end{figure}

We finally analysed the trajectories traced by corner events labelled as independent motion, grouped according to the ground truth (Fig.~\ref{fig:tracking}). Only trajectories along the $x$ sensor plane are shown for clarity. Ideally, all corner events within the ground truth region should be classified as independent motion (i.e. in Fig.~\ref{fig:trackX}), and all corner events from the background should be ego-motion (i.e. in Fig.~\ref{fig:trackY}). Despite the recall of $\sim 40~\%$ a consistent detection is still achieved \textit{over time}, indicating a segmentation algorithm could potentially achieve a consistent result. Importantly, false positives are sparse and don't form a coherent pattern, such that a simple filter could easily reject such detections of independent motion. 

\section{Conclusions} \label{sec:conclusion}

In this work, we have presented an event-based independent motion detector using the event camera, which disentangles the independent motion that occurs in the visual scene from the robot ego-motion. As background clutter can induce many additional events (but irrelevant for certain tasks), this task is crucial for event-driven scenarios where cameras are non stationary (on a robot). The use of cluster events reduces the data flow to the most informative events, enabling efficient, real-time implementation of many different event-driven vision algorithms for robotics. 
We detect and track corners in the space of events and learn the correlation between their motion and robot's joint velocities, when there is no moving object in the scene. We then label as belonging to independent motion corner events whose motion does not agree with the predicted velocity. 

We model ego-motion with first-order statistics, relying on the assumption of negligible motion parallax (which depends on the structure of the scene) and motion induced by rotation around the optical axis of the cameras (the head roll), as in~\cite{Fanello13icra,Kumar15humanoids}. This assumption did not affect the result as the algorithm was able to detect independent motion, with a precision of $\sim 90\%$, consistently with changing speed of both the target and the head. However we plan to model and learn the ego-motion using an affine motion model to be robust to head rotations.
The detection can be problematic when the object changes direction, as the relative motion with the background approaches zero. However we do not need dense detection in time as sparse detections of independent motion can be used as triggering locations for visual tracking.
Finally we show that sparse optical flow can be effectively used to address independent motion detection, reducing therefore the amount of data to process.

We plan to use these sparse detections to segment a moving object in a cluttered scene on the iCub, implementing an event-based attention mechanism driven by the motion of the target, which would facilitate visual tracking.    

\section*{Acknowledgment}
This research was supported by the Swiss National Science Foundation
through the National Center of Competence in Research Robotics.



\bibliographystyle{IEEEtran}
\bibliography{references}

\end{document}